# Facial transformations of ancient portraits: the face of Caesar

**Amelia Carolina Sparavigna**
Department of Applied Science and Technology, Politecnico di Torino, Italy

*Some software solutions used to obtain the facial transformations can help investigating the artistic metamorphosis of the ancient portraits of the same person. An analysis with a freely available software of portraitures of Julius Caesar is proposed, showing his several "morphs". The software helps enhancing the mood the artist added to a portrait.*

In image processing, a facial transformation can be used to simulate rejuvenation or aging of faces, and even to produce variations, the morphs, from a face into another, often through the smooth transitions of a metamorphic sequence. However, if we consider the representation of a face in painting, photography or sculpture, here too we can find a sort of morph, in particular when the face expression is the predominant feature of the portraiture. In fact, the intent of a portrait is not only to display the face, but also to show the mood of the person. As a result, as told in [1], for some viewers a portrait is a sort of living being: "Portraits contain images that with bewildering success pretend not to be signs or tokens invented by artists, but rather aim to represent the manner in which their subjects would appear to the viewer in life".

In ancient times, the sculptures in marble or stone and casts in bronze were the means to preserve the likenesses of men of distinction [2]. Portraits on substrates having lesser durability and resistance such as on wood and canvas existed, however they rarely survived, such as some mummy portraits of the Egyptian Ptolemaic Dynasty [3]. In some cases, for instance for the busts of Julius Caesar, we have the possibility to investigate several portraits, or "morphs", of the same person and determine how the variations imposed by each artist determined a shift from an appearance to another.

A first detailed study, accompanied by photographic pictures, of the marble busts of Julius Caesar was performed in 1903 by the architect Frank J. Scott. As he tells in the introduction of his book [2] on the portraitures of Caesar, in Rome, "for a few centuries before and after Julius Caesar, nearly every citizen who acquired reputation or notoriety in Rome was likely to be well or badly portrayed by some sculptor". For his critical study and comparison, Scott decided to see personally all Caesar busts and statues and "to obtain photographs, drawings, or casts of every marble or bronze that could be found assuming to be of him." To prepare himself for the task, he became acquainted with the writings of archaeologists and iconographists, in particular he used the studies of J.J. Bernoulli of the University of Basle, and his Romische Ikonographie published in 1882.

Scott's book contains the descriptions of busts known at his times. Some busts therefore are not included in it. One is that known as the Tusculum bust. It is considered the oldest bust of Caesar [4-6]. As told in [7], this bust "has been identified as that of Caesar by the typical saddle of the crown accentuated by the bold forehead, the angular jawline, the 'vulturine' neck, and —last but not least— by the ironic lines of the mouth. Here is the same vision and will, but the Clementia (mercy) is more concealed by a stronger sense of irony. One can see the ruler of the world advancing, and sense there is a claiming of ownership, and an inaccessibility. … In fact, this head could have been fashioned up to two years before (some coins) were struck, because in the time between 46 and 44 BC a number of statues were consecrated to Caesar in Italy. In this case too, the sculptor for whom Caesar posed was obviously not unimpressed. This marble head itself became a model for later statues during the time of the emperors". The Tusculum bust was then the model for a colossal head in the Farnese collection and from this head of several other statues.

Reference 7 continues telling that the head from Tusculum originated a conjecture regarding another head, that of the Torlonia museum, "which fascinates the researchers and leaves them divided: namely, is this his real face, or the face that met the expectations of the time? The features are the same except for the direction of the eyebrows, but the expression is completely different. This head seems to have had some influence on the later statues of Caesar, ones in which the Clementia was accentuated as it is with the head in the Vatican." [7] It is then clear that the "morphing" in the creations of artists was common and used to have the desired mood rendered in the portrait.

In 2007, another bust was discovered in the Rhone river near Arles, and proposed as a portrait of Caesar. In [8], the Arles bust is compared with the Tusculum and Farnese heads. There is a certain agreement in the morphs, however, some scholars disagree that this is a portrait of Julius Caesar [9]. A web site, Ref.10, is offering a rich collection of pictures of portraits and here we find also the Pantelleria bust, found in 2006, a bust showing the face of a young man.

To compare all the busts of Caesar it would be better to possess their 3D scanning; however, we can try a different approach based on a face recognition software, such as those used for investigations. In [11], I applied a software freely available at the web site in20years.com [12], which allows to see how a person looks after several years. Using a picture showing a full face, it is possible to select the gender and a possible use of drugs, and to predict what the face would look like in 20 or 30 years. Nothing else is required and therefore the result is not influenced by user's preferences.

In [11], I used in20years.com software to compare the busts of the Roman emperor Augustus, portrayed when he was young and as a man, having results which are good and realistic. Here I am proposing to use it to compare the portraitures of Julius Caesar. In fact, the main result of the software for aging faces on the images of busts, is to give them a realistic appearance, and, probably due to the aging effect, to enhance the mood in the portrait.

In the Figure 1, we can see twenty portraitures, obtained applying in20years.com software to the images of ancient marble busts. Here the list of them:

01 - Arles discovered in September–October 2007 in the Rhone River near Arles, France.
02 – Marble, ca. 40 BC, Rome, Torlonia Museum.
03 - 1st quarter of the 1st century A.D., Corinth, Archaeological Museum.
04 – Marble, Age of August (27 BC — AD 14), Pisa, Campo Santo.
05 - White marble, late 1st — early 2nd century AD, Turin, Museum of Antiquities.
06 - White marble, Woburn Abbey.
07 - Farnese head of Julius Caesar from the Trajan's forum, marble, AD 117—138, Naples, National Archaeological Museum.
08 - Age of Julii-Claudii, white marble, Munich, Residenz.
09 – Farnese head, time of Trajan, Naples, National Archaeological Museum.
10 - Julian-Claudian copy of the portrait of the 1st century BC, Rome, Vatican Museums.
11 - White marble, Vienna, Museum of Art History.
12 - White marble, time of Trajan, Parma, National Museum of Antiquities.
13 - Fragment of the statue, Rome, Capitoline Museums.
14 – Marble, Age of August, Rome, Vatican Museums.
15 - White marble, time of Augustus, Rome, Vatican Museums.
16 - White marble, Time of Augustus, Rome, Vatican Museums.
17 - Found in Pantelleria (Sicily), 2006, dated to the middle of the first century CE.
18 - The Tusculum bust, fine-grained white marble, 45—43 BC, Turin, Museum of Antiquities.
19 - Plaster cast from the portrait head of Julius Caesar in the Museum of Antiquities, Turin.
20 - Private collection, Rome, [13,14].

Of course, some of the portraitures after processing display a little bit old face, however the features of the faces are realistic. The Tusculum portraiture is N.18 and 19 of the Figure 1. Another bust, N.20, has a rather faint image of Caesar, but the use of software [10] enhances it so that it is clear the bust is linked with the Tusculum bust. In N.17, there is the Pantelleria bust, which is the portrait of a young man. After aging the face, we can compare it with the Tusculum bust, and it seems that a good agreement exists. N.16 is the Vatican bust, and it is clear the accentuated Clementia [7], in particular when compared with the Tusculum, where "the Clementia is more concealed by a stronger sense of irony".

The Figure 1 is then a collection of a "morphing" that artists produced over a period of two centuries in their representation of Julius Caesar, where the mood of each face is enhanced by the use of a software giving them a natural feature. Let us conclude with N.1, the Arles bust: the processing seems to give us the picture of a living person. Below, the detail of Figure 1, showing on the right the bust, where I have restored the damaged nose, and on the left the image after processing with in20years software. Besides the good processing result, it is also enhanced the agreement with the Torlonia bust, the N.2 in Figure 1. Therefore, it seems that is the aging effect which enhances the mood of the portrait.

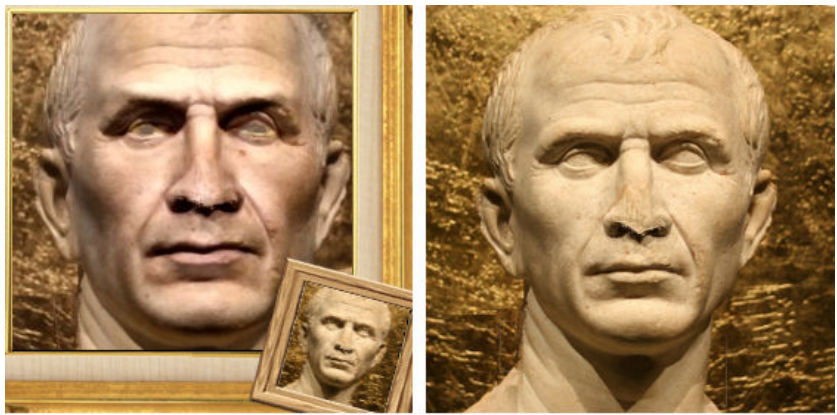

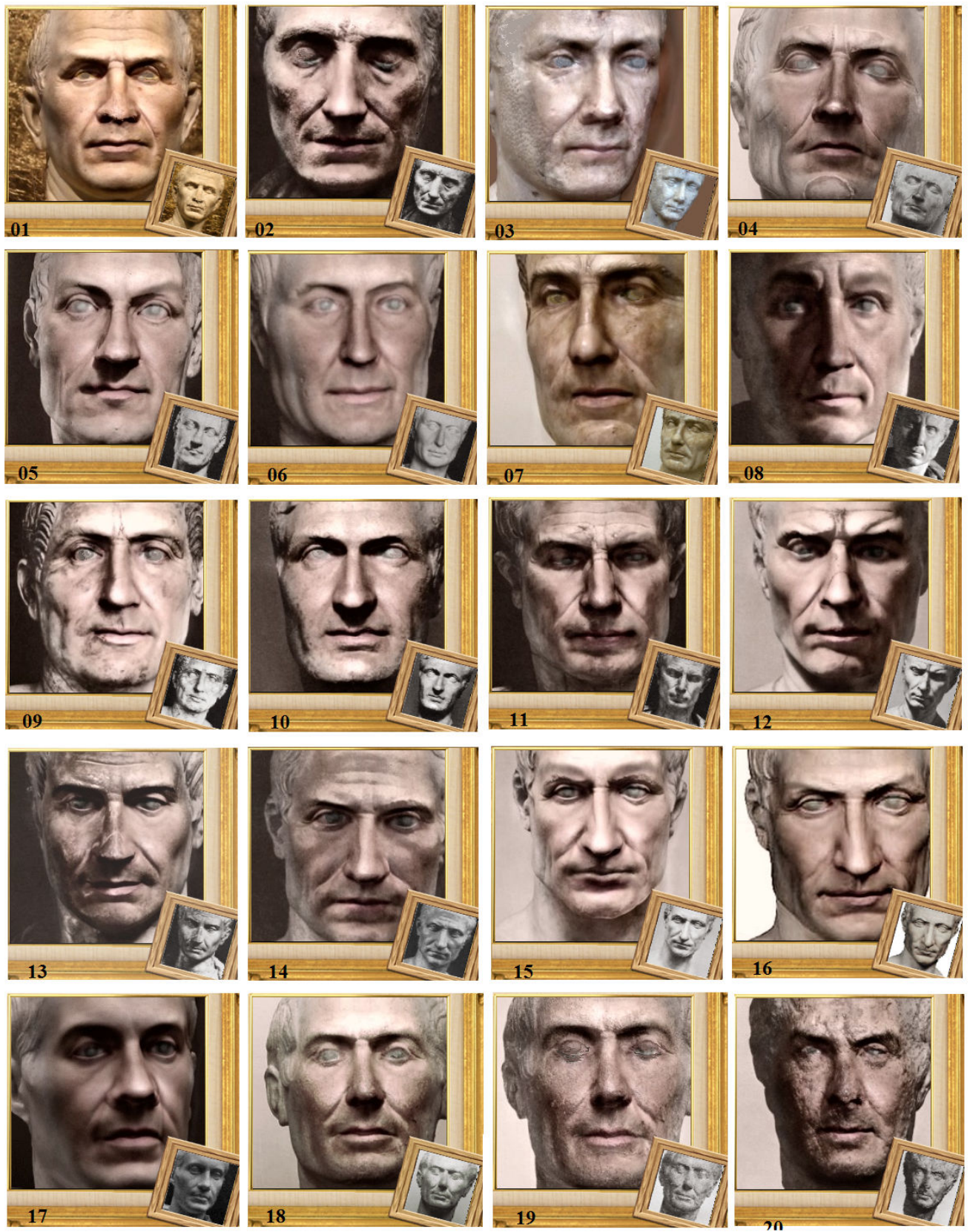

Figure 1 – Portraitures of Julius Caesar, obtained applying in20years.com software to the images of ancient marble busts. N.1 is the Arles bust, N.2 the Torlonia bust, N.9 Farnese bust, N.17 Pantelleria, N.18 and 19 the Tusculum bust.